\documentclass[times,referee,twocolumn,final,authoryear]{elsarticle}

\usepackage{ycviu}
\usepackage{framed,multirow}

\usepackage{amssymb}
\usepackage{latexsym}

\usepackage{url}
\usepackage[dvipsnames]{xcolor}
\definecolor{newcolor}{rgb}{.8,.349,.1}

\newcommand{\Edit}[1]{\textcolor{black}{#1}}

\usepackage{tabularx}
\usepackage{cite}
\usepackage{balance}
\usepackage{subfigure}
\usepackage{enumitem}
\setlist{nosep}
\usepackage{booktabs}
\usepackage[multiple]{footmisc}

\usepackage{float}
\usepackage{stfloats}

\usepackage{multirow}
\usepackage{hyperref}

\journal{Computer Vision and Image Understanding}

\begin{document}

\thispagestyle{empty}

\ifpreprint
  \setcounter{page}{1}
\else
  \setcounter{page}{1}
\fi

\begin{frontmatter}

\title{Anabranch Network for Camouflaged Object Segmentation}

\author[1]{Trung-Nghia \snm{Le}\corref{cor1}} 
\cortext[cor1]{Trung-Nghia Le (ltnghia@its.iis.u-tokyo.ac.jp) contributed to this work while he interned at University of Dayton. He is currently a postdoc at the Institute of Industrial Science, the University of Tokyo, Japan.
Tam V. Nguyen (tamnguyen@udayton.edu) is the corresponding author.
}
\author[2]{Tam V. \snm{Nguyen}\corref{cor1}}
\author[2]{Zhongliang \snm{Nie}}
\author[3]{Minh-Triet \snm{Tran}}
\author[4]{Akihiro \snm{Sugimoto}}

\address[1]{Department of Informatics, The Graduate University for Advanced Studies (SOKENDAI), Tokyo, Japan}
\address[2]{Department of Computer Science, University of Dayton, Ohio, 45469, United States of America}
\address[3]{University of Science, VNU-HCM, Ho Chi Minh, Vietnam}
\address[4]{National Institute of Informatics, Tokyo, Japan}

\received{1 May 2013}
\finalform{10 May 2013}
\accepted{13 May 2013}
\availableonline{15 May 2013}
\communicated{S. Sarkar}

\begin{abstract}

Camouflaged objects attempt to conceal their texture into the background and discriminating them from the background is hard even for human beings. The main objective of this paper is to explore the camouflaged object segmentation problem, namely, segmenting the camouflaged object(s) for a given image. This problem has not been well studied in spite of a wide range of potential applications including the preservation of wild animals and the discovery of new species, surveillance systems, search-and-rescue missions in the event of natural disasters such as earthquakes, floods or hurricanes. This paper addresses a new challenging problem of camouflaged object segmentation. To address this problem, we provide a new image dataset of camouflaged objects for benchmarking purposes. In addition, we propose a general end-to-end network, called the Anabranch Network, that leverages both classification and segmentation tasks. Different from existing networks for segmentation, our proposed network possesses the second branch for classification to predict the probability of containing camouflaged object(s) in an image, which is then fused into the main branch for segmentation to boost up the segmentation accuracy. Extensive experiments conducted on the newly built dataset demonstrate the effectiveness of our network using various fully convolutional networks.

\end{abstract}


\end{frontmatter}


\section{Introduction}
\label{sec:intro}

\begin{figure*}[t]
    \centering
        \includegraphics[width=1\textwidth]{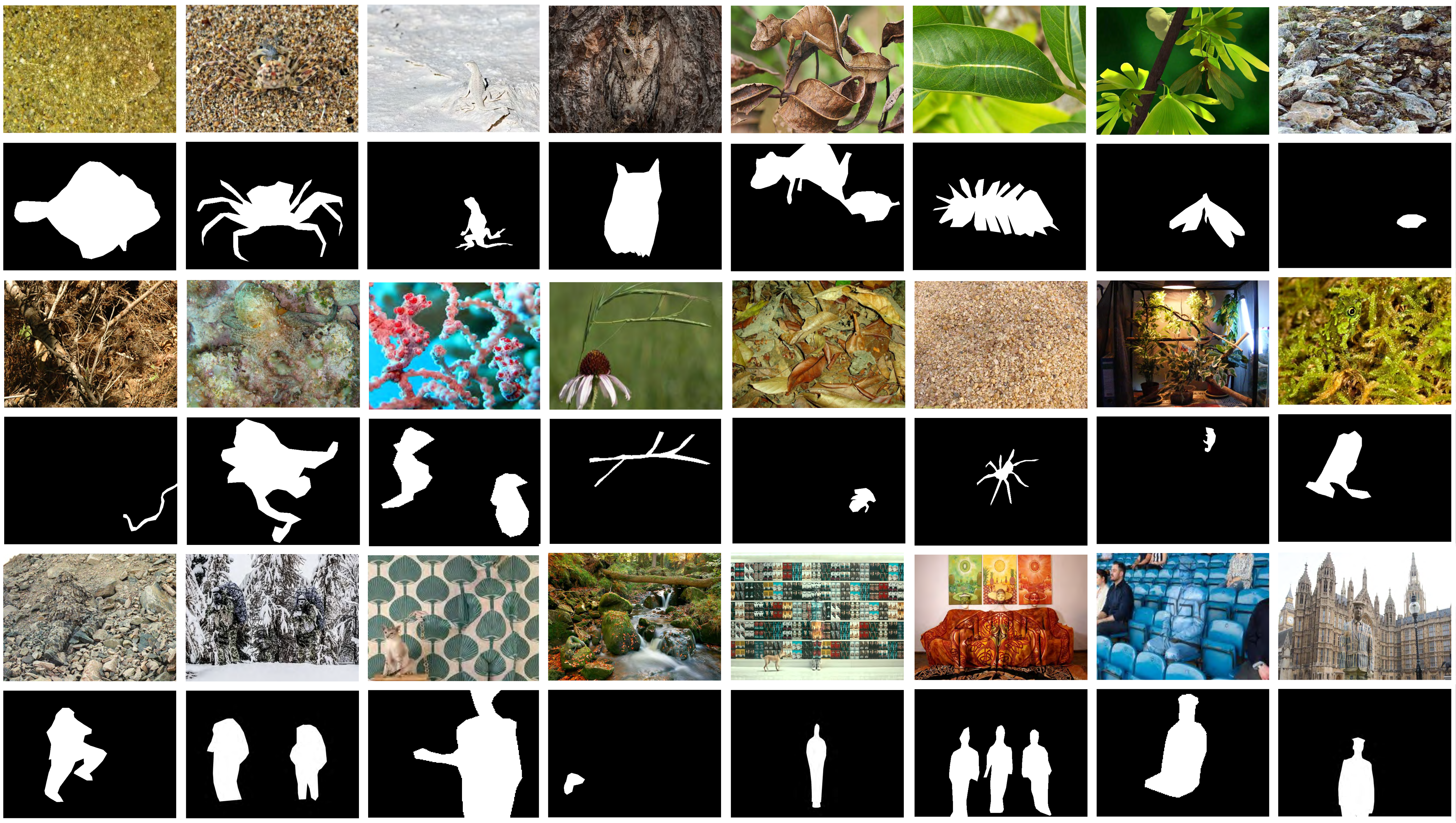}
    \caption{\Edit{A few examples from our Camouflaged Object (CAMO) dataset with corresponding pixel-level annotations. Camouflaged objects attempt to conceal their texture into the background.}}
    \label{fig:CAMO}
\end{figure*}



Camouflage is an attempt to conceal the texture of a foreground object into the background~\citep{Sujit-ICEECS2013}. The term ``\textit{camouflage}'' was first coined from nature where animals used to hide themselves from predators by changing their body pattern, texture, or color. Unlike salient objects, camouflaged objects are hard to detect in their nature even for human beings. Autonomously detecting/segmenting camouflaged objects is thus a challenging task where discriminative features do not play an important role anymore since we have to ignore objects that capture our attention. While detecting camouflaged objects is technically difficult on one hand, it is beneficial in various practical scenarios, on the other hand, which include surveillance systems, and search-and-rescue missions. For example, detecting and segmenting camouflaged objects in images definitely helps us search for camouflaged animals for the preservation of wild animals and the discovery of new species. Nevertheless, camouflaged object segmentation has not been well explored in the literature.

There exist two types of camouflaged objects in reality: naturally camouflaged objects such as animals or insects hiding themselves from their predators, and artificially camouflaged objects such as soldiers and weapons disguised by artificial texture patterns. These objects first evaluate their surrounding environment and then change to camouflaged textures via the colorization of cloths or cover. For either type of camouflaged objects, it is not obvious to identify them in images.
Figure \ref{fig:CAMO} shows a few examples of camouflaged objects in real life. Based on this figure, we can easily see how challenging camouflaged object segmentation is.



Camouflaged objects have naturally evolved to exploit weaknesses in the visual system of their prey or predator, and thus understanding such a mechanism will provide some insights for segmenting camouflaged objects. Without any prior, we human beings easily miss detecting camouflaged objects; however, once we are informed a camouflaged object exists in an image, we can carefully scan the entire image to detect it. This also comes up in nature: once the predator has an awareness of camouflaged animals in the scene, then it makes efforts to localize them for hunting. Therefore, utilizing the awareness as a prior can be an essential cue for camouflaged object segmentation. Then, the challenge is what awareness should be incorporated in what way. We consider that classification scheme can be used as the awareness of existing of camouflaged objects in an image in order to combine with detection/segmentation scheme.

As reviewed in \Edit{Singh \textit{et al.}~\citep{Sujit-ICEECS2013}}, there are a few methods that study camouflaged object detection in simple contexts. These methods use hand-crafted low-level features such as color, edge or texture to detect camouflaged objects. Such hand-crafted low-level features are designed to be as much discriminative as possible for detecting and segmenting objects, which is the opposite way to camouflage. Therefore, their performances are limited to camouflaged object segmentation. Furthermore, these methods can be applied only to relatively low-resolution images with a uniform background. 
Deep features, on the other hand, are known to outperform low-level features in many tasks in computer vision such as image classification~\citep{Krizhevsky-NIPS2012}, image semantic
segmentation~\citep{Shelhamer-PAMI2016}, action recognition~\citep{Tran-ICCV2015}, and face recognition~\citep{Wen-ECCV2016}. 
We thus expect that deep features can replace low-level features even for camouflaged object segmentation by incorporating them into a new network. To use them, however, a large number of data of camouflaged objects are required.  Nevertheless, there is no public dataset (both for training data and testing data) for the camouflaged object segmentation problem. This obviously poses serious problems to (1) train deep learning models and (2) evaluate the performance of the proposed network. 

The overall contribution of this paper is two-fold: 
\begin{itemize}

\item We provide a new image dataset 
of camouflaged objects to promote new methods for camouflaged object segmentation. Our newly constructed \textbf{Cam}ouflaged \textbf{O}bject (CAMO) dataset consists of 1250 images, each of which contains at least one camouflaged object. Pixel-wise ground-truths are manually annotated to each image. Furthermore, images in the CAMO dataset involve a variety of challenging scenarios such as \textit{object appearance, background clutter, shape complexity, small object, object occlusion, multiple objects,} and \textit{distraction}. We emphasize that this is the very first dataset for camouflage segmentation.

\item We propose a novel end-to-end network, called the Anabranch\footnote{\href{https://en.wikipedia.org/wiki/Anabranch}{An anabranch is a section of a river or stream that diverts from the main channel or stem of the watercourse and rejoins the main stem downstream.}} Network (ANet), which is \textit{general, conceptually simple, flexible, and efficient} for camouflaged object segmentation. Our proposed ANet leverages the advantages of different network streams, namely, the classification stream and the segmentation stream, in order to segment camouflaged objects. The classification stream is used as the awareness of existing of camouflaged object(s) in an image. This design is motivated by the observation that \textit{there is no guarantee that a camouflaged object is always present in an image} and thus we first identify whether a camouflaged object is present in an image and then only if exists, the object(s) should be accurately segmented; nothing is segmented otherwise. Extensive experiments verify the effectiveness of our proposed ANet when the network is applied on various fully convolutional networks (FCNs) for camouflaged object segmentation.
\end{itemize}

\Edit{The datasets, evaluation scripts, models, and results are publicly available at our websites \footnote{\href{https://sites.google.com/view/ltnghia/research/camo}{https://sites.google.com/view/ltnghia/research/camo}}\textsuperscript{,}\footnote{\href{https://sites.google.com/site/vantam/research}{https://sites.google.com/site/vantam/research}}}

The remainder of this paper is organized as follows. Section~\ref{sec:related_work} summarizes the related work. Next, Section~\ref{sec:dataset} and Section~\ref{section:framework} introduce the constructed dataset and the proposed network, respectively. Section~\ref{section:experiments} then reports extensive experiments over the proposed method and baselines on the newly constructed dataset. Section~\ref{joint_training} discusses the joint training of ANet. Finally, Section~\ref{sec:conclusion} draws the conclusion and paves the way for the future work.

\section{Related Work}
\label{sec:related_work}

This section first reviews existing works on camouflaged object segmentation and salient object segmentation, which share some similarities. We also clarify that the camouflaged object segmentation is more challenging than the salient object segmentation. Then, we briefly introduce some advancements of two-stream networks in several computer vision tasks.

\subsection{Camouflaged Object Segmentation}

As addressed in the intensive survey compiled by Singh \textit{et al.}~\citep{Sujit-ICEECS2013}, most of the related work \citep{Kavitha-IJEST2011, Pan-MAS2011,Siricharoen-ICGCS2010,Song-ICMT2010, Yin-PE2011} uses handcrafted, low-level features for camouflaged region detection/segmentation where the similarity between the camouflaged region and the background is evaluated using low-level features, \textit{e.g.}, color, shape, orientation, and brightness~\citep{Galun-ICCV2003, Song-ICMT2010, Xue-MTA2016}. We note that there is little work that directly deals with camouflaged object detection; most of the work is dedicated to detecting the foreground region even when some of its texture is similar to the background. Pan \textit{et al.}~\citep{Pan-MAS2011} proposed a 3D convexity based method to detect camouflaged objects in images. Liu \textit{et al.}~\citep{Liu-TIP2012} integrated the top-down information based on the expectation maximization framework for foreground object detection. Sengottuvelan \textit{et al.}~\citep{Sengottuvelan-ICETET2008} applied a gray level co-occurrence matrix method to identify the camouflaged object in images with a simple background. In the case where some parts of a moving object and the background share the similar texture, optical flow is combined with color to detect/segment the moving object~\citep{Yin-PE2011, Gallego-ICIP2014}. All these methods use handcrafted low-level features and work for only a few cases where images/videos have the simple and non-uniform background. Their performances are also unsatisfactory in camouflage detection/segmentation when there is a strong similarity between the foreground and the background. 

One of the main reasons why there has been very little work on camouflaged object segmentation is the lack of the standard dataset for this problem. A benchmarking dataset is thus mandatory to develop advanced techniques for camouflaged object segmentation. Note that the dataset has to include both training data and testing data. To our best of knowledge, there is so far no standard dataset for camouflage object segmentation, which has the sufficient number of data for training and evaluating deep networks. A few of datasets related to camouflaged animals have been proposed, but they have a limited number of samples. Bideau \textit{et al.}~\citep{Bideau-ECCV2016} proposed a camouflaged animal video dataset but the dataset has only 9 videos, and camouflaged animals really exist in a third of videos. The unpublished dataset proposed by Skurowski et.al.\footnote{\href{http://zgwisk.aei.polsl.pl/index.php/en/research/other-research/63-animal-camouflage-analysis}{http://zgwisk.aei.polsl.pl/index.php/en/research/other-research/63-animal-camouflage-analysis}} has only 76 images of camouflaged animals. Therefore, we collect a new camouflaged object dataset. Note that our work is the first work solve the real camouflage segmentation problem in diverse images. Different from the two above datasets, our constructed dataset consists of both camouflaged animals and human.

\subsection{Salient Object Segmentation}

Salient object predictors aim to detect and segment salient objects in images. Though ``saliency'' is opposed to ``camouflage'', techniques developed for salient object segmentation may be useful for camouflaged object segmentation. This is because the two tasks highlight image regions with certain characteristics.

Early work on salient object segmentation is based on biologically-inspired approaches~\citep{Itti-PAMI1998, Koch-MI1987} where used features are the contrast of low-level features such as orientation of edges, or direction of movement. Since human vision is sensitive to color, different approaches using color were proposed where the contrast of color features is locally or globally analyzed~\citep{Achanta-CVPR2009, Cheng-CVPR2011}. Local analysis methods estimate the saliency of a particular image region against its neighborhoods based on color histogram comparison~\citep{Cheng-CVPR2011}; global analysis methods achieve globally consistent results by computing color dissimilarities to the mean image color~\citep{Achanta-CVPR2009}. Various patch-based methods were also proposed that estimate dissimilarity between image patches~\citep{Cheng-CVPR2011, Margolin-CVPR2013, Zhang-ICCV2015}. 
To leverage the advantage of deep learning, recent methods first conduct superpixel segmentation and then feed segmented regions into a convolutional neural network (CNN) individually to obtain saliency scores~\citep{ltnghia-TIP2018, Li-CVPR2015, LWang-CVPR2015}.  More recent methods modified fully convolutional networks (FCNs) to compute a pixel-wise saliency map~\citep{Hou-CVPR2017, ltnghia-BMVC2017, Li-CVPR2017, Liu-CVPR2016, Wang-ICCV2017, Wang-CVPR2017}. Skip-layer structures are usually employed in FCNs to obtain multi-scale feature maps, which are critically needed to segment a salient object~\citep{ltnghia-BMVC2017, Liu-CVPR2016, Hou-CVPR2017}. Li \textit{et al.}~\citep{Li-CVPR2016} modified the FCN to combine pixel-wise and segment-wise feature maps for saliency value prediction. Wang \textit{et al.}~\citep{Wang-ICCV2017} integrated a pyramid pooling module and a multi-stage refinement mechanism into an FCN for salient object segmentation. Wang \textit{et al.}~\citep{Wang-CVPR2017} proposed a two-stage training method to compute saliency maps using image-level tags. \Edit{Object proposal based networks, which are originally developed for category-specific object detection, are also utilized to identify salient objects~\citep{Han-SPM2018, ltnghia-WACV2019}. In addition, salient objects from common categories can be co-segmented by exploiting correspondence relationship among multiple relevant images~\citep{Han-TIP2018, Yao-TIP2017}.}

\begin{figure}[t]
    \centering
    \begin{tabularx}{\linewidth}{*{3}{X}}
        \hfill\includegraphics[width=0.9\linewidth]{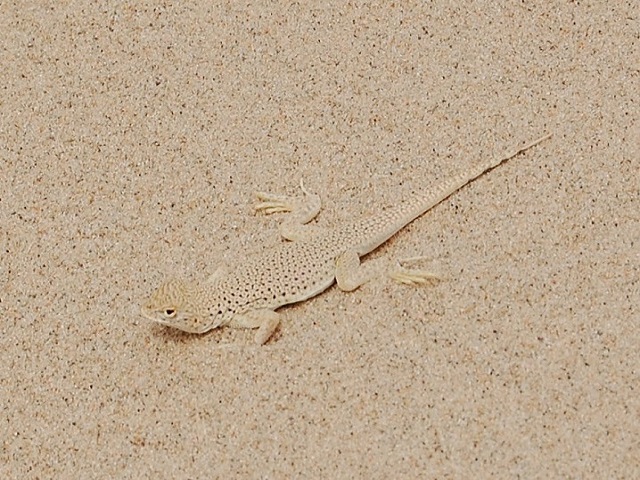}\hspace*{\fill} & 
        \hfill\includegraphics[width=1\linewidth]{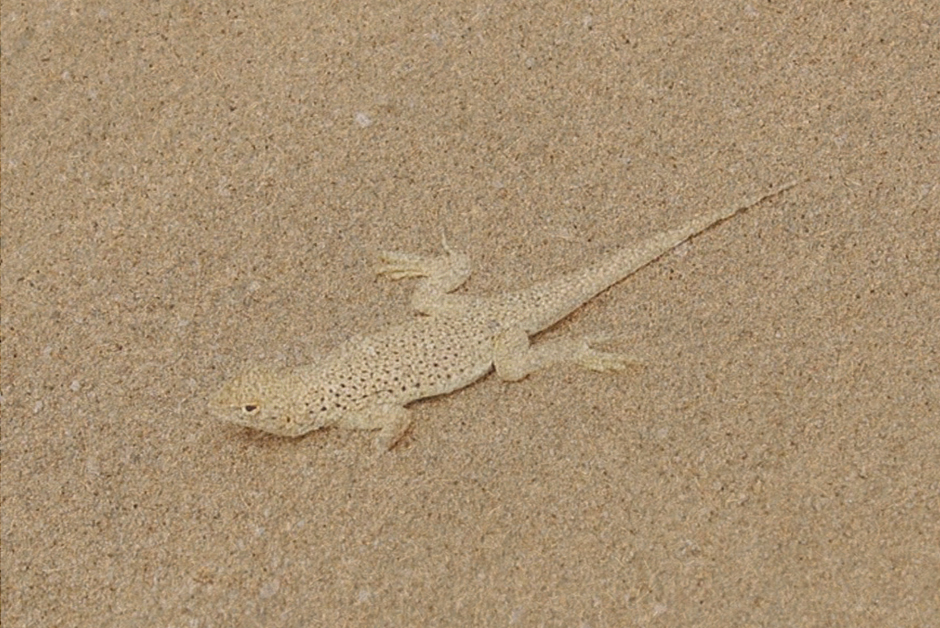}\hspace*{\fill} & 
        \hfill\includegraphics[width=1\linewidth]{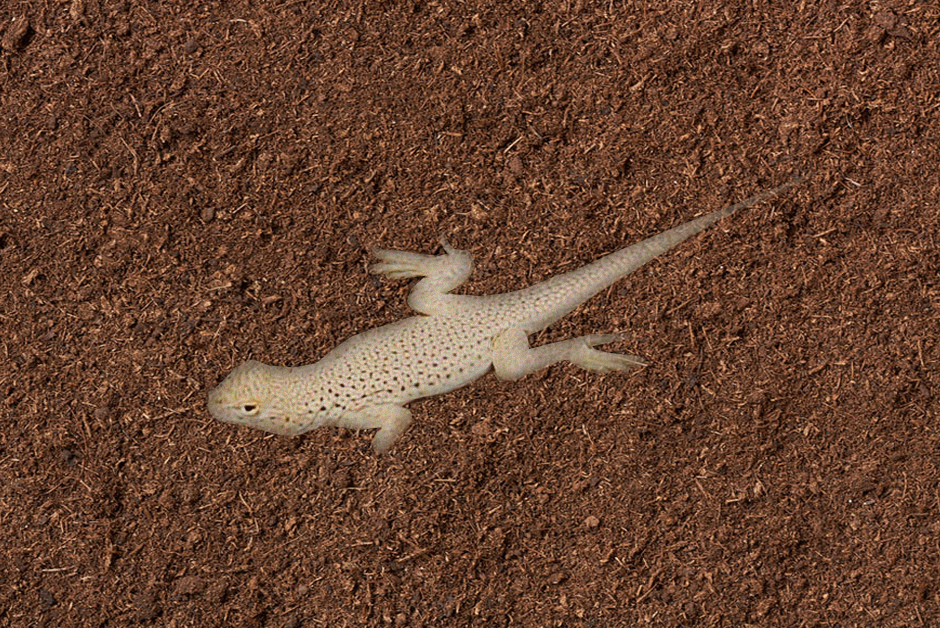}\hspace*{\fill} \\
        \centering \footnotesize{(a) Camouflaged object} &
        \centering \footnotesize{(b) Ambiguous} &
        \centering \footnotesize{(c) Salient object} 
    \end{tabularx}
    \caption{Ambiguous differences between camouflaged objects and salient objects.}
    \label{fig:camouflage_saliency}
\end{figure}

It is worth noting that camouflaged object segmentation is more challenging than salient object segmentation. In fact, salient objects tend to be outstanding and discriminative from the background. We thus only need to focus on identifying such outstanding and discriminative regions to segment salient objects. Camouflaged objects, however, tend to conceal themselves into the background environment by decreasing discriminativeness as much as possible. This makes it hard to identify boundaries of camouflaged objects. In particular, when the background is cluttered, camouflaged objects blend with the background too much, and discriminating them from the background becomes even harder. 


We remark that the same object can be either a camouflaged object or a salient object, meaning that the differences between camouflaged objects and salient objects are ambiguous (cf. Fig.~\ref{fig:camouflage_saliency}). 
This suggests that some salient object detection methods may work for the camouflaged object segmentation task; however, this is not the case. Indeed, existing methods on salient object segmentation are designed to segment objects for any reason because a salient object is assumed to always exist in a training/testing image. On the contrary, there is no guarantee that a camouflaged object is always present in an image. The approach to camouflaged object segmentation should be completely different so that it allows us to handle practical scenarios where we do not know whether or not a target object exists.

\subsection{Two-Stream Network}

There are several works that use two-stream architectures~\citep{Simonyan-NIPS2014,  Feichtenhofer-NIPS2016,  Limin-ECCV2016, Jain-CVPR2017, Kaiming-ICCV2017, Yi-CVPR2017}. 
Fusion networks were proposed for action recognition~\citep{Simonyan-NIPS2014, Feichtenhofer-NIPS2016, Limin-ECCV2016} and object segmentation~\citep{Jain-CVPR2017} in videos. The network consists of an appearance stream and a motion stream, and the final result is fused from the outputs of the two streams. 
Two streams, however, have the same architecture for the same task despite the fact that they receive different inputs (\textit{i.e.}, video frame and optical flow). Two streams (\textit{i.e.}, a classification stream and a segmentation stream) of region-based networks for instance segmentation~\citep{Kaiming-ICCV2017, Yi-CVPR2017} have different architectures for different tasks. The two streams are built up from the region proposal network~\citep{Ren-NIPS2015} and then work on specific regions.

In contrast with the above mentioned two-stream networks, the two streams (the classification stream and the segmentation stream) of our proposed ANet have different architectures mutually reinforce each other on the whole image for different tasks. The output of the classification stream is fused with that of the segmentation stream to enhance segmentation result.  In this sense, the role of the classification stream is complementary to that of the segmentation stream.

\section{Camouflaged Object Dataset}
\label{sec:dataset}

\subsection{Motivation}

In the literature, no dataset is available for camouflaged object segmentation. To promote camouflaged object segmentation, a publicly available dataset with pixel-wise ground-truth annotation is mandatory. Note that the dataset has to include both training data and testing data. The training data is suitable for training a deep neural network whereas the testing data is used for evaluation. Therefore, we aim to construct a dataset, \textbf{Cam}ouflaged \textbf{O}bject (\textbf{CAMO}) dataset, to promote advancements in camouflaged object segmentation and its evaluation. We stress that our CAMO dataset is the very first dataset for camouflaged object segmentation.

\subsection{Dataset Construction}

To build the dataset, we initially collected 3000 images in which at least one camouflaged object exists. We collected the images of camouflaged objects from the Internet with keywords: ``camouflaged objects'', ``camouflaged animals'', ``concealed objects'', ``hidden animals'', ``camouflaged soldier'', ``human body painting'', \textit{etc}. Note that we consider both types of camouflaged objects, namely, natural and artificial camouflage as mentioned in Section~\ref{sec:intro}. Then, we manually discarded images with low resolution or duplication. Finally, we ended up with 1250 images. To avoid inconsistency in labeling ground-truth, we asked three people to annotate camouflaged objects in all 1250 images individually using a custom designed interactive segmentation tool. On average, each person takes 2-5 minutes to annotate one image depending on its complexity. The annotation stage spanned about one month. 

\subsection{Dataset Description}

\begin{figure}[t]
    \centering
        \includegraphics[width=1\linewidth]{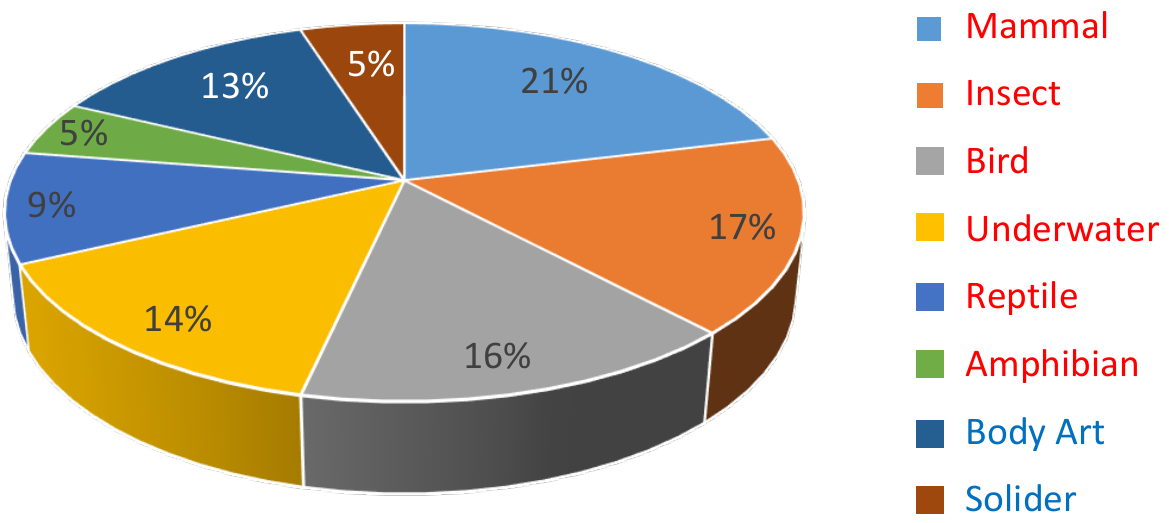} 
    \caption{Categories in Camouflaged Object Dataset. Natural camouflaged objects are shown in \textcolor{red}{red} while artificial camouflaged objects are shown in \textcolor{blue}{blue}.}    
    \label{fig:CAMO_categories}
\end{figure}

\begin{figure}[t]
    \centering
        \includegraphics[width=1\linewidth]{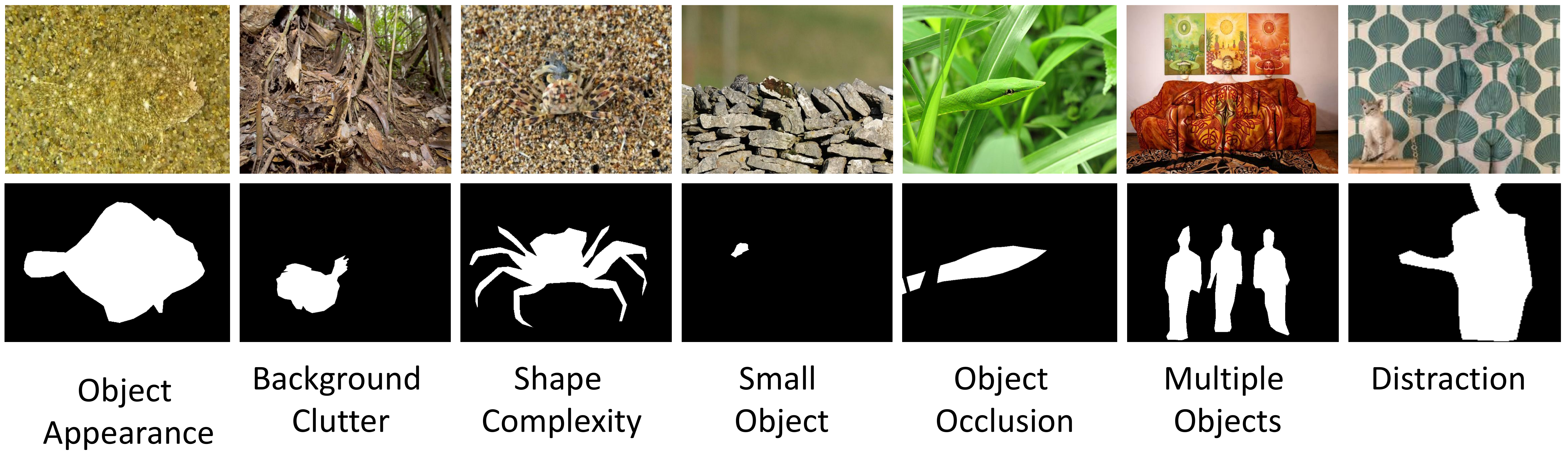} 
        \centering \footnotesize{(a) Examples of challenging attributes} 
        
        \includegraphics[width=1\linewidth]{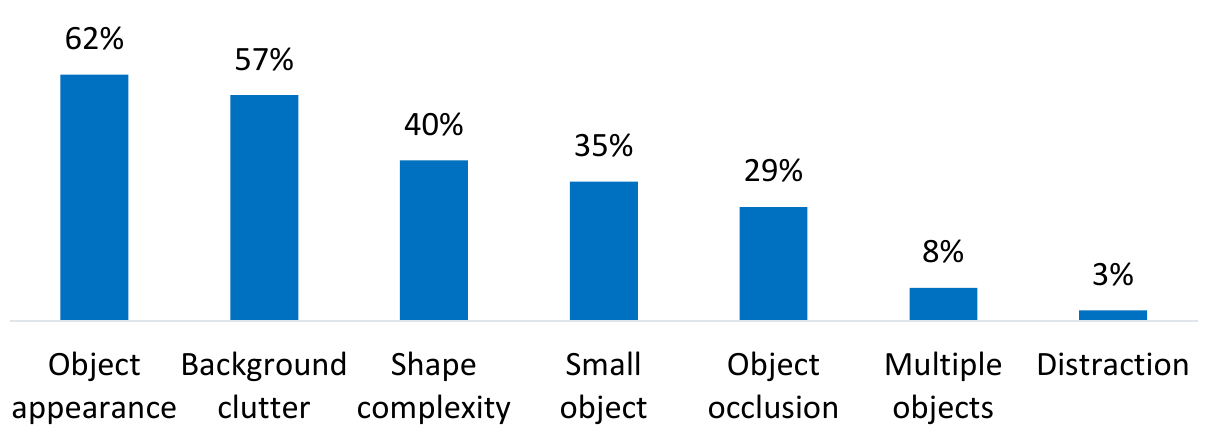} 
        \centering \footnotesize{(b) Attribute distribution over the CAMO dataset} 
    \caption{Some examples of challenging attributes and the attribute distribution over the CAMO dataset.}
\label{fig:attributes}
\end{figure}

We describe in this section the Camouflaged Object (CAMO) dataset specifically designed for the task of camouflaged object segmentation. Some examples are shown in Fig.~\ref{fig:CAMO} with the corresponding ground-truth label annotations. We focus on two categories, \textit{\textit{i.e.}}, naturally camouflaged objects and artificially camouflaged objects, which usually correspond to animals and humans in the real world, respectively. Camouflaged animals consist of amphibians, birds, insects, mammals, reptiles, and underwater animals in various environments, \textit{\textit{i.e.},} ground, underwater, desert, forest, mountain, and snow. Camouflaged human falls into soldiers on the battlefields and human body painting arts. The ratios for each category are shown in Fig.~\ref{fig:CAMO_categories}.

In our CAMO dataset, it is also noteworthy that multiple objects, including separate single objects and spatially connected/overlapping objects, possibly exist in some images ($8\%$). This also makes our dataset more challenging for camouflaged object segmentation. The challenge of the CAMO dataset is also enhanced due to some attributes, \textit{\textit{i.e.}} object appearance, background clutter, shape complexity, small object, object occlusion, and distraction (cf. Fig.~\ref{fig:attributes}):
\begin{itemize}
    \item \textbf{Object appearance}: The object has a similar color appearance with the background, causing large ambiguity in segmentation.
    \item \textbf{Background clutter}: The background is not uniform but contains small-scale structures or is composed of several complex parts.
    \item \textbf{Shape complexity}: The object has complex boundaries such as thin parts and holes, which is usually the case for the legs of insects.
    \item \textbf{Small object}: The ratio between the camouflaged object area and the whole image area is smaller than 0.1.
    \item \textbf{Object occlusion}: The object is occluded, resulting in disconnected parts or in touching the image border. 
    \item \textbf{Distraction}: The image contains distracted objects, resulting in losing attention to camouflaged objects. This makes camouflaged objects more difficult to discover even by human beings.
\end{itemize}


\section{Anabranch Network} 
\label{section:framework}

In this section, we give a detailed description of our Anabranch Network (\textbf{ANet}). Specifically, we first discuss the motivation of network design. Then, we introduce the architecture of ANet.

\subsection{Network Design}


As mentioned above, there is no guarantee that a camouflaged object is always present in the scene. 
Therefore, a method that systematically segments objects for any image will not work.
Moreover, directly applying discriminative features from segmentation models (\textit{i.e.}, semantic segmentation and salient object segmentation, \textit{etc.}) to camouflaged object segmentation is not effective because camouflaged objects conceal their texture into the surrounding environment. 
In order to segment camouflaged objects, we need an additional function that identifies whether a camouflaged object exists in an image.
To do so, each pixel needs to be classified as a part of camouflaged objects or not.
Such classification can be used not only to enhance segmentation accuracy but also to segment multiple camouflaged objects.
Indeed, this classification only strengthens features extracted from the camouflaged part and weakens features extracted from a non-camouflaged part. 
Accordingly, segmentation and classification tasks should be closely combined in the network with different architectures for camouflaged object segmentation.


\subsection{Network Architecture}

\begin{figure*}[t]
    \centering
        \includegraphics[width=1\textwidth]{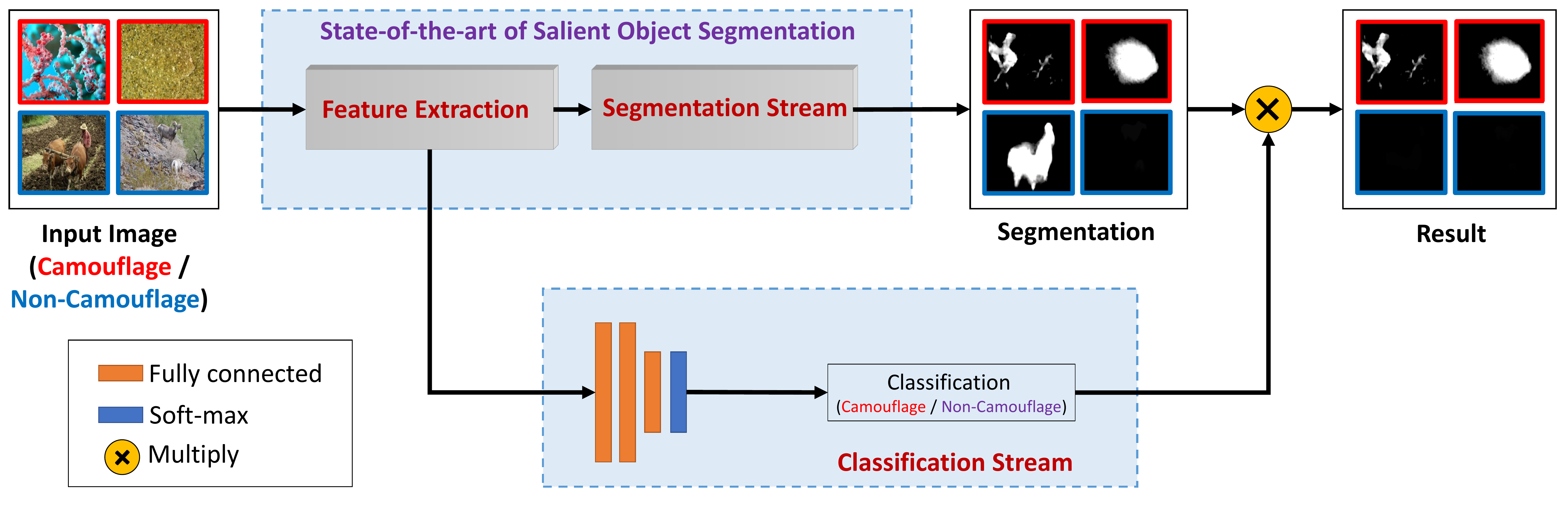}
    \caption{Overview of our proposed Anabranch Network (ANet). The proposed network leverages the strength of both image classification and semantic segmentation tasks for camouflaged object segmentation.}
    \label{fig:overview}
\end{figure*}

Figure \ref{fig:overview} depicts the overview of our proposed ANet, a general network for camouflaged object segmentation. The ANet leverages the advantages of the two different network models, namely, the classification network model based on a convolutional neural network (CNN) and the segmentation network model based on a fully convolutional network (FCN). The CNN reveals object hypotheses in the image~\citep{Bolei}, yielding a cue on whether or not the input image contains camouflaged object(s). We thus aim to train the CNN model so that it classifies two classes: camouflaged image class and non-camouflaged image one. The FCN, on the other hand, provides us with the pixel-wise semantic information in the image. We consider the output of the FCN as the semantic information of different objects in the image. The two outputs from the two network streams are finally fused to produce a pixel-wisely accurate camouflage map which uniformly covers camouflaged objects. We note that our classification stream output, \textit{\textit{i.e.}}, the probability (scalar), is multiplied ($\otimes$) with each pixel value of 2D map produced by the segmentation stream. 


\textbf{Segmentation Stream:} The input and the output of the segmentation stream are both images, and, thus, any end-to-end FCN for segmentation can be employed.
In the sense that we can employ any FCN for segmentation, ANet can be considered as a general network.

Though, as addressed above, methods for salient object detection alone may not work for accurate camouflaged object segmentation, effective compensation by the classification stream can be expected to boost up the performance in the ANet.
Recent state-of-the-art methods for salient object detection have demonstrated their good performances.
In addition, as we observed in Fig.~\ref{fig:camouflage_saliency}, we have ambiguity in difference between camouflaged objects and salient objects.
Based on these reasons, we consider a saliency model is suitable to use into ANet.
We can easily swap the model with any other saliency model.
We remark that it is also interesting to utilize saliency models in a similar but different domain such as camouflaged object segmentation.


\begin{figure*}[!t]
    \centering
        \includegraphics[width=1\linewidth]{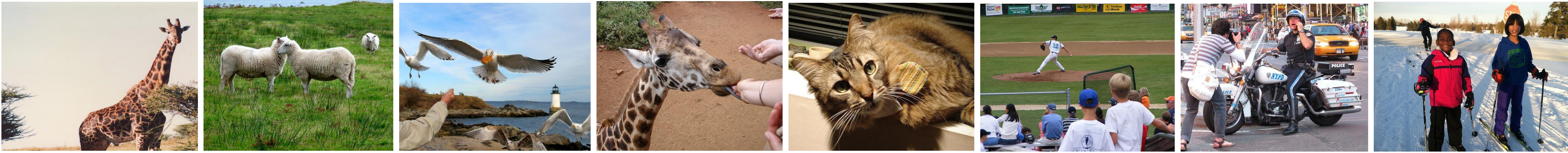}
    \caption{Examples of non-camouflaged objects from MS-COCO.}
    \label{fig:noncamouflage}
\end{figure*}

\textbf{Classification Stream:}
We build the classification stream on top of convolution layers, which play as feature extraction module, of FCNs. In particular, we use three fully connected layers with 4096, 4096, and 2 filters, respectively (cf. Fig.~\ref{fig:overview} and Table~\ref{tab:cls_branch}). We note that each of the first two fully connected layers is followed by a Rectified Linear Unit (ReLU) activation layer~\citep{Krizhevsky-NIPS2012}. The last layer is the soft-max layer. To train this stream, we use soft-max with the cross-entropy loss. During the training process, we use dropout layers with a dropout rate of $50\%$ after the first two fully connected layers to avoid over-fitting.

\begin{table}[t]
\centering
\caption{Architecture of the classification stream.}
\label{tab:cls_branch}
\small
\begin{tabular}{llccc}
\toprule
\textbf{No} & \textbf{Layer} & \textbf{Output} & \textbf{Dropping Rate} \\ \midrule
1 & Fully connected & 2048 & \\
2 & ReLU & 2048 & \\ 
3 & Dropout & 2048 & 50\% \\ 
4 & Fully connected & 2048 & \\ 
5 & ReLU & 2048 & \\ 
6 & Dropout & 2048 & 50\% \\ 
7 & Fully connected & 2 & \\ 
8 & Soft-max & 2 & \\ \bottomrule
\end{tabular}
\end{table}

As seen above, our introduced ANet is conceptually simple. 
Each stream in ANet has its own task and; thus, each stream is expected to be individually trained using data suitable for the task
to enhance its ability.
Then, the fusion of the two streams into one boosts up segmentation accuracy of camouflaged objects. 
Furthermore, ANet is flexible in the sense that we can employ any end-to-end FCN and easily switch it with another. 
As we see in our experimental results, ANet is robust since it 
maintains good segmentation accuracy in both cases of with and without a camouflaged object in an input image.
We also see its computational efficiency.

\section{Experiments} 
\label{section:experiments}

In this section, we first introduce evaluation datasets and evaluation criteria used in experiments. Then we describe implementing details as well as network training of our proposed ANet and other baselines. In experiments, we show the potential ability of transforming domain from salient object segmentation to camouflaged object segmentation. We next compare instances of our ANet with the FCN models fine-tuned on two evaluation datasets, to demonstrate that classifying camouflaged objects and non-camouflaged objects can boost up the camouflaged object segmentation. We also present the efficiency of our general ANet through short network training and fast running time. These results can be considered as the first baselines for the camouflaged object segmentation problem. Finally, we discuss the challenge of the proposed CAMO dataset to show rooms for further research on this problem.

\begin{table}[t]
\centering
\caption{CAMO-COCO dataset used in our experiments.}
\label{tab:dataset}
\scriptsize
\begin{tabular}{l|ccc}
\toprule
\textbf{} & \multicolumn{1}{l}{\textbf{Training}} & \multicolumn{1}{l}{\textbf{Testing}} & \multicolumn{1}{l}{\textbf{Total}} \\ \midrule
\textbf{Camouflaged object images (CAMO)} & 1000 & 250 & 1250 \\ 
\textbf{Non-Camouflaged object images (MS-COCO)} & 1000 & 250 & 1250 \\ \midrule
\textbf{Total} & 2000 & 500 & 2500 \\ \bottomrule
\end{tabular}
\end{table}

\subsection{Datasets and Experimentation Setup}\label{section:dataset}

Any image in the CAMO dataset contains at least one camouflaged object. There is thus prior information that camouflaged objects are always present. In a realistic scenario, however, there is no guarantee that a camouflaged object is always present in an image. 
Therefore, we set up another dataset, called CAMO-COCO\footnote{The link to the dataset will be available along with the publication of this paper.}, consisting of camouflaged object images and non-camouflaged object images. 
We used the entire images in the CAMO dataset for the camouflaged object images and collected additional 1250 images from the MS-COCO dataset~\citep{Lin-ECCV2014} for the non-camouflaged object images (cf. Fig.~\ref{fig:noncamouflage}). 
We note that we created zero-mask ground-truth labels (all pixels have zero values) for the non-camouflaged object images. 
For each of the camouflaged image and non-camouflaged object image sets, we randomly chose 80\% for training (1000 images) and used the remaining 20\% for testing (250 images). 
Table~\ref{tab:dataset} shows the number of images in the CAMO-COCO dataset used  
for experiments.  

\subsection{Evaluation Criteria} \label{section:metrics}

We used the F-measure ($F_\beta$)~\citep{Achanta-CVPR2009}, Intersection Over Union (IOU)~\citep{Long-ICCV2015}, and Mean Absolute Error (MAE)
as the metrics to evaluate obtained results. 
The first metric, F-measure, is a balanced measurement between precision and recall as follows: 
\begin{equation}
{F_\beta } = \frac{{\left( {1 + {\beta ^2}} \right)Precision \times Recall}}{{{\beta ^2} \times Precision + Recall}}.
\end{equation}
Note that we set $\beta^2=0.3$ as used in \citep{Achanta-CVPR2009} to put an emphasis on precision. 
IOU is the area ratio of the overlapping against the union between the predicted camouflage map and the ground-truth map. Meanwhile, MAE is the average of the pixel-wise absolute differences between the predicted camouflage map 
and the ground-truth. %


For MAE, we used the raw grayscale camouflage map. For the other metrics, we binarized the results depending on two contexts. In the first context, we assume that camouflaged objects are always present in every image like salient objects; we used an adaptive threshold~\citep{Jia-ICCV2013} $\theta=\mu+\eta$ where $\mu$ and $\eta$ are the mean value and the standard deviation of the map, respectively. In the second context which is much closer to a real-world scenario, we assume that the existence of camouflaged objects is not guaranteed in each image; we used the fixed threshold $\theta=0.5$.


\subsection{Implementation Details} \label{sec:implement}

To demonstrate the generality and flexibility of our proposed ANet, we employ various recent state-of-the-art FCN-based salient object segmentation models for the segmentation stream.  They are DHS~\citep{Liu-CVPR2016}, DSS~\citep{Hou-CVPR2017},  SRM~\citep{Wang-ICCV2017}, and WSS\citep{Wang-CVPR2017}.

For each employed model, we train two streams sequentially. We first fine-tuned the segmentation stream using the CAMO dataset from the publicly available pre-trained model and then trained the classification stream using CAMO-COCO with fixed parameters in the segmentation stream. The final model is considered as our baseline for camouflaged object segmentation.


In both training steps, we set the size of each mini-batch to 2, and used the Stochastic Gradient Descent (SGD) optimization~\citep{Rumelhart-Neurocomputing1988} with a moment $\beta=0.9$ and a weight decay of 0.0005. 
We trained the segmentation stream for 10 epochs (corresponding to 5k iterations) with the learning rate of $10^{-4}$ and trained the classification stream for 3 epochs (3k iterations) with the learning rate of $10^{-6}$. During the fine-tuning process, a simple data augmentation technique was used to avoid over-fitting. Images were rescaled to the resolution of the specific FCN employed in the segmentation stream and randomly flipped horizontally.

We note that for comparison, our employed FCN models (original) were trained on CAMO (or CAMO-COCO) dataset for 10 epochs with the learning rate of $10^{-4}$ and the other parameters were set similarly to our ANet.
We also remark that we implemented our method in C/C++, using Caffe~\citep{Jia-MM2014} toolbox and conducted all the experiments on a computer with a Core i7 3.6 GHz processor, 32 GB of RAM, and two GTX 1080 GPUs.

\subsection{Experimental Results}

\subsubsection{Baseline Evaluation}
\begin{table*}[t]
\centering
\caption{Experimental results on two datasets: CAMO dataset (the left part), and CAMO-COCO dataset (the right part). The evaluation is based on F-measure~\citep{Achanta-CVPR2009} (the higher the better), IOU~\citep{Long-ICCV2015} (the higher the better), and MAE (the smaller the better). The 1st and 2nd places are shown in \textcolor{blue}{\textbf{blue}} and \textcolor{red}{\textbf{red}}, respectively.}
\label{tab:result}
\resizebox{1\textwidth}{!}{%

\begin{tabular}{|l|c|cc|cc|l|c|cc|cc|}
\toprule
 \textbf{Dataset in test} & \multicolumn{5}{c|}{\textbf{CAMO}} &  & \multicolumn{5}{c|}{\textbf{CAMO-COCO}} \\ \cline{1-6} \cline{8-12} 
\textbf{Method} & \textbf{} & \multicolumn{2}{c|}{\textbf{Adaptive Threshold}} & \multicolumn{2}{c|}{\textbf{Fixed Threshold}} &  & \textbf{} & \multicolumn{2}{c|}{\textbf{Adaptive Threshold}} & \multicolumn{2}{c|}{\textbf{Fixed Threshold}} \\
  & \textbf{MAE $\Downarrow$} & \textbf{F$_\beta$ $\Uparrow$} & \textbf{IOU $\Uparrow$} & \textbf{F$_\beta$ $\Uparrow$} & \textbf{IOU $\Uparrow$} &  & \textbf{MAE $\Downarrow$} & \textbf{F$_\beta$ $\Uparrow$} & \textbf{IOU $\Uparrow$} & \textbf{F$_\beta$ $\Uparrow$} & \textbf{IOU $\Uparrow$} \\ \cmidrule{1-6} \cmidrule{8-12} 
 
DHS~\citep{Liu-CVPR2016} (pre-trained) & 0.173 & 0.548 & 0.351 & 0.562 & 0.332 &  & 0.204 & 0.746 & 0.571 & 0.750 & 0.549 \\
DHS (fine-tuned with CAMO) & \textcolor{blue}{\textbf{0.129}} & \textcolor{blue}{\textbf{0.640}} & \textcolor{blue}{\textbf{0.459}} & \textcolor{blue}{\textbf{0.643}} & \textcolor{blue}{\textbf{0.444}} &  & \textcolor{red}{\textbf{0.169}} & 0.794 & 0.633 & 0.793 & 0.618 \\
DHS (fine-tuned with CAMO-COCO) & 0.138 & 0.596 & 0.388 & 0.614 & 0.367 &  & \textcolor{blue}{\textbf{0.072}} & \textcolor{red}{\textbf{0.796}} & \textcolor{red}{\textbf{0.679}} & \textcolor{red}{\textbf{0.808}} & \textcolor{red}{\textbf{0.681}} \\
ANet-DHS (baseline) & \textcolor{red}{\textbf{0.130}} & \textcolor{red}{\textbf{0.626}} & \textcolor{red}{\textbf{0.437}} & \textcolor{red}{\textbf{0.631}} & \textcolor{red}{\textbf{0.423}} &  & \textcolor{blue}{\textbf{0.072}} & \textcolor{blue}{\textbf{0.812}} & \textcolor{blue}{\textbf{0.712}} & \textcolor{blue}{\textbf{0.814}} & \textcolor{blue}{\textbf{0.705}} \\
\midrule 

DSS~\citep{Hou-CVPR2017} (pre-trained) & 0.157 & 0.564 & 0.320 & 0.563 & 0.333 &  & 0.176 & 0.757 & 0.559 & 0.756 & 0.570 \\
DSS (fine-tuned with CAMO) & \textcolor{red}{\textbf{0.141}} & \textcolor{blue}{\textbf{0.622}} & \textcolor{blue}{\textbf{0.425}} & \textcolor{blue}{\textbf{0.631}} & \textcolor{blue}{\textbf{0.420}} &  & 0.152 & \textcolor{red}{\textbf{0.791}} & 0.633 & \textcolor{red}{\textbf{0.795}} & 0.630 \\
DSS (fine-tuned with CAMO-COCO) & 0.145 & 0.582 & 0.385 & 0.584 & 0.381 &  & \textcolor{red}{\textbf{0.076}} & 0.790 & \textcolor{red}{\textbf{0.686}} & 0.792 & \textcolor{red}{\textbf{0.687}} \\
ANet-DSS (baseline) & \textcolor{blue}{\textbf{0.132}} & \textcolor{red}{\textbf{0.587}} & \textcolor{red}{\textbf{0.404}} & \textcolor{red}{\textbf{0.607}} & \textcolor{red}{\textbf{0.390}} &  & \textcolor{blue}{\textbf{0.067}} & \textcolor{blue}{\textbf{0.795}} & \textcolor{blue}{\textbf{0.701}} & \textcolor{blue}{\textbf{0.804}} & \textcolor{blue}{\textbf{0.694}} \\
\midrule 

SRM~\citep{Wang-ICCV2017} (pre-trained) & 0.171 & 0.448 & 0.258 & 0.425 & 0.213 &  & 0.191 & 0.699 & 0.535 & 0.685 & 0.502 \\
SRM (fine-tuned with CAMO) & \textcolor{blue}{\textbf{0.120}} & \textcolor{blue}{\textbf{0.683}} & \textcolor{blue}{\textbf{0.507}} & \textcolor{blue}{\textbf{0.688}} & \textcolor{blue}{\textbf{0.498}} &  & 0.176 & 0.815 & 0.651 & 0.812 & 0.634 \\
SRM (fine-tuned with CAMO-COCO) & 0.127 & \textcolor{red}{\textbf{0.663}} & 0.454 & 0.656 & 0.421 &  & \textcolor{blue}{\textbf{0.067}} & \textcolor{blue}{\textbf{0.830}} & \textcolor{red}{\textbf{0.717}} & \textcolor{blue}{\textbf{0.831}} & \textcolor{red}{\textbf{0.708}} \\
ANet-SRM (baseline) & \textcolor{red}{\textbf{0.126}} & 0.654 & \textcolor{red}{\textbf{0.475}} & \textcolor{red}{\textbf{0.662}} & \textcolor{red}{\textbf{0.466}} &  & \textcolor{red}{\textbf{0.069}} & \textcolor{red}{\textbf{0.826}} & \textcolor{blue}{\textbf{0.732}} & \textcolor{red}{\textbf{0.830}} & \textcolor{blue}{\textbf{0.727}} \\
\midrule 

WSS~\citep{Wang-CVPR2017} (pre-trained) & 0.178 & 0.559 & 0.323 & 0.531 & 0.265 &  & 0.197 & 0.754 & 0.567 & 0.740 & 0.528 \\ 
WSS (fine-tuned with CAMO) & \textcolor{red}{\textbf{0.145}} & \textcolor{red}{\textbf{0.658}} & \textcolor{blue}{\textbf{0.477}} & \textcolor{blue}{\textbf{0.661}} & \textcolor{blue}{\textbf{0.465}} &  & 0.174 & 0.807 & 0.649 & \textcolor{red}{\textbf{0.810}} & 0.637 \\
WSS (fine-tuned with CAMO-COCO) & 0.149 & 0.642 & 0.439 & 0.638 & 0.382 &  & \textcolor{red}{\textbf{0.085}} & \textcolor{red}{\textbf{0.811}} & \textcolor{red}{\textbf{0.678}} & \textcolor{blue}{\textbf{0.820}} & \textcolor{red}{\textbf{0.687}} \\
ANet-WSS (baseline) & \textcolor{blue}{\textbf{0.140}} & \textcolor{blue}{\textbf{0.661}} & \textcolor{red}{\textbf{0.459}} & \textcolor{red}{\textbf{0.643}} & \textcolor{red}{\textbf{0.407}} &  & \textcolor{blue}{\textbf{0.078}} & \textcolor{blue}{\textbf{0.826}} & \textcolor{blue}{\textbf{0.710}} & \textcolor{blue}{\textbf{0.820}} & \textcolor{blue}{\textbf{0.697}} \\
\bottomrule
\end{tabular}
}
\end{table*}

\begin{figure*}[bp]
    \centering
    \includegraphics[width=1\textwidth]{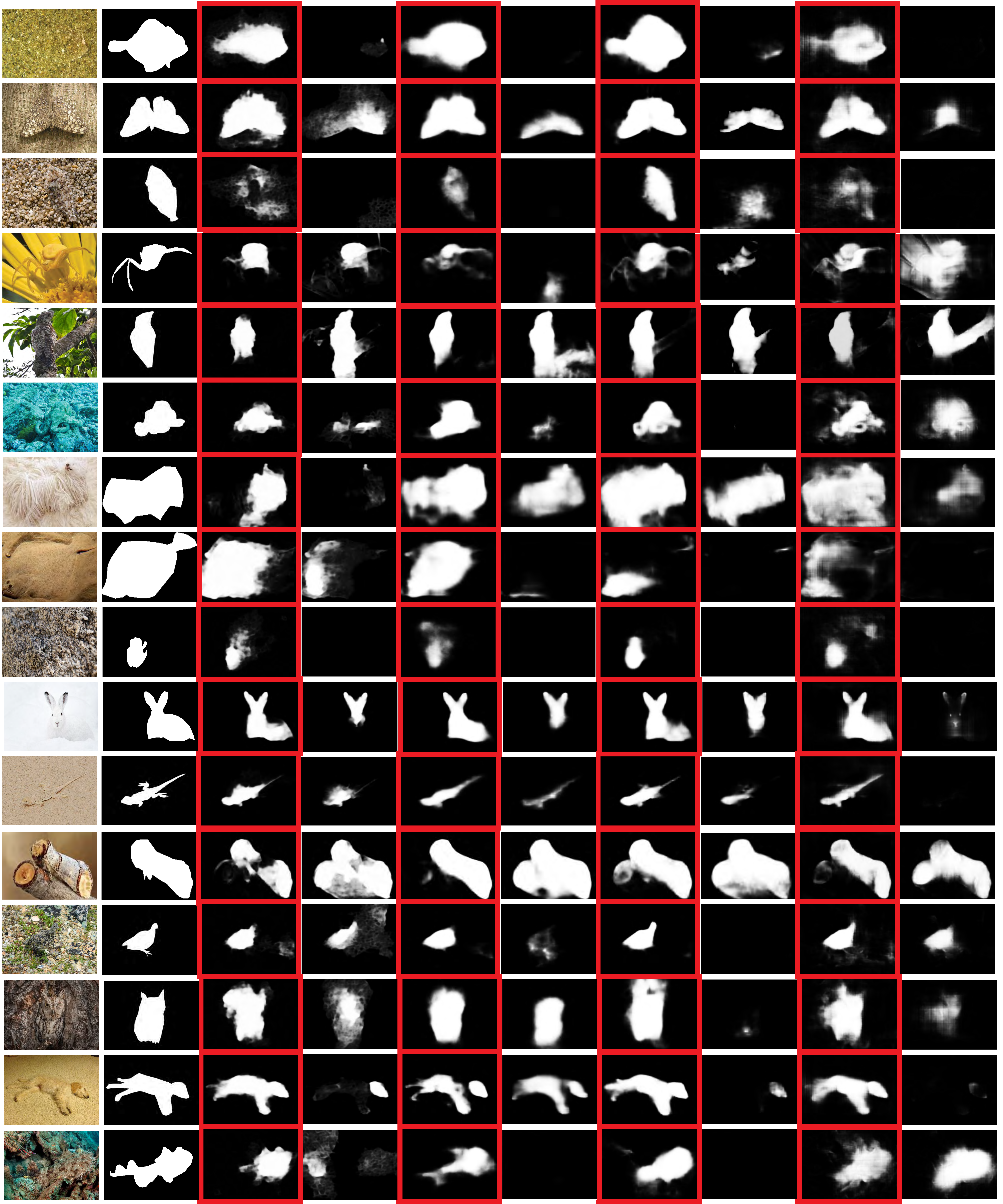}
    \caption{\Edit{Visual comparison of our ANet employing different FCNs on natural camouflaged object images (CAMO dataset). From left to right, input image and ground-truth are followed by outputs obtained using ANet with DHS~\citep{Liu-CVPR2016}, DHS (pre-trained), ANet with DSS~\citep{Hou-CVPR2017}, DSS (pre-trained), ANet with SRM~\citep{Wang-ICCV2017}, SRM (pre-trained), ANet with WSS~\citep{Wang-CVPR2017}, and WSS (pre-trained) in this order. The results obtained by ANet is surrounded with red rectangles.}}
\label{fig:visualization1}
\end{figure*}

\begin{figure*}[bp]
    \centering
    \includegraphics[width=1\textwidth]{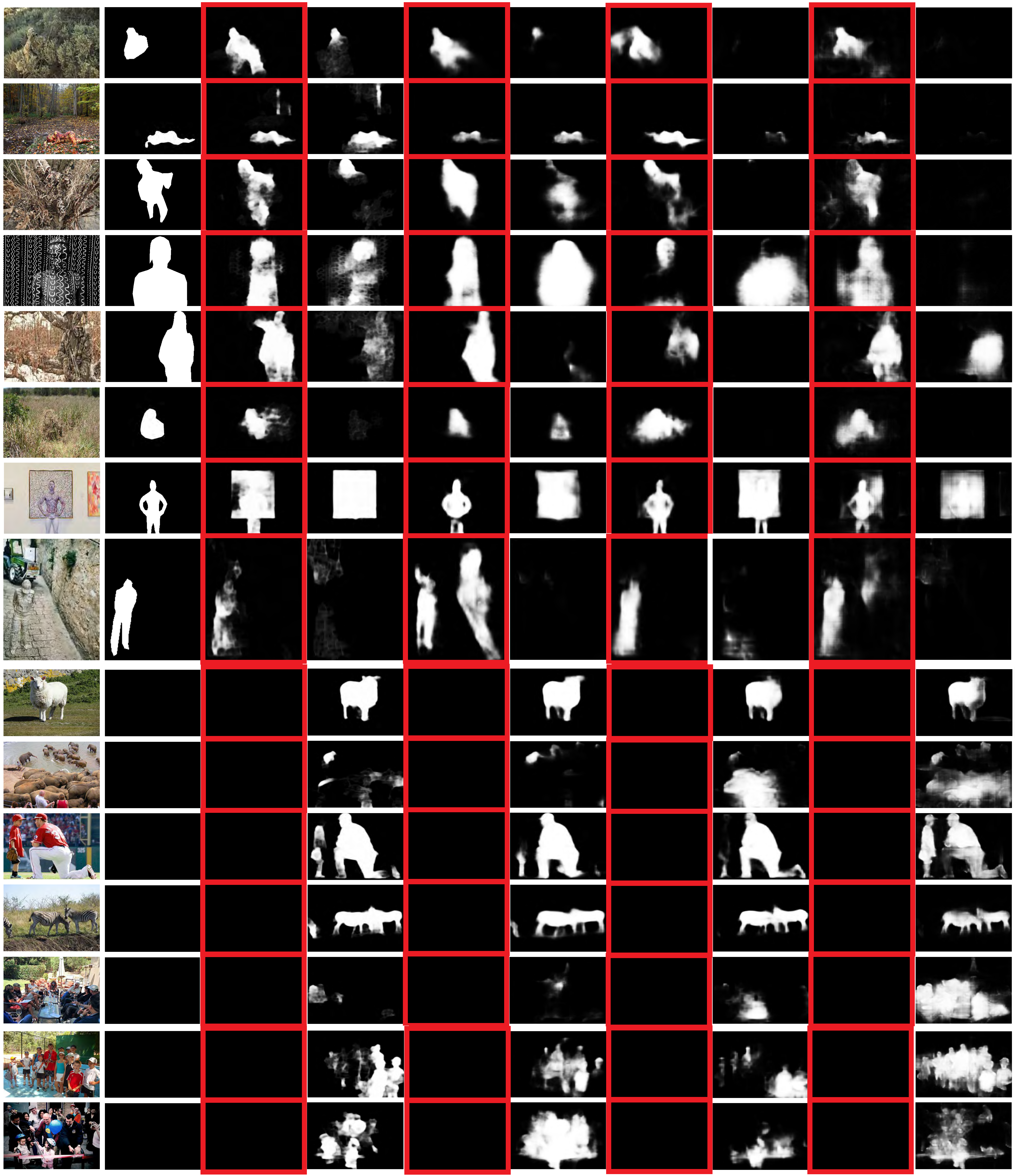}
    \caption{\Edit{Visual comparison of our ANet employing different FCNs. From left to right, input image and ground-truth are followed by outputs obtained using ANet with DHS~\citep{Liu-CVPR2016}, DHS (pre-trained), ANet with DSS~\citep{Hou-CVPR2017}, DSS (pre-trained), ANet with SRM~\citep{Wang-ICCV2017}, SRM (pre-trained), ANet with WSS~\citep{Wang-CVPR2017}, and WSS (pre-trained) in this order. The first eight rows are images with artificial camouflaged objects (CAMO dataset), and the last seven rows are images without camouflaged objects (COCO dataset). The results obtained by ANet is surrounded with red rectangles.}}
\label{fig:visualization2}
\end{figure*}

We evaluated our model (baseline) on two datasets: CAMO (camouflaged object images only) and CAMO-COCO (both camouflaged and non-camouflaged object images) where only test images in each dataset were used for this evaluation.  We employed an FCN (DHS~\citep{Liu-CVPR2016}, DSS~\citep{Hou-CVPR2017},  SRM~\citep{Wang-ICCV2017}, or WSS\citep{Wang-CVPR2017}) as the segmentation stream. For comparison, we also evaluated each of used FCNs to see how it works for camouflaged object segmentation. Each was evaluated with three different models: the pre-trained models on saliency datasets (\textit{e.g.} MSRA~\citep{Cheng-PAMI2015}, DUT-OMRON~\citep{Yang-CVPR2013}, and DUTS~\citep{Wang-CVPR2017}), the models that were fine-tuned with CAMO or CAMO-COCO (the model is specified by the word inside the parentheses). \Edit{Figures \ref{fig:visualization1}, \ref{fig:visualization2},} and Table~\ref{tab:result} illustrate the obtained results of our experiments. We note that ANet-DHS (baseline), for example, denotes our proposed model employing DHS as the segmentation stream.

\Edit{Figures \ref{fig:visualization1} and \ref{fig:visualization2} show that ANet yields better results than its employed original FCN (pre-trained) independently of the choice of the FCN.}  In particular, for non-camouflaged object images, the superiority of performances of ANet against its original FCN is distinguished.  This suggests that the fusion with the classification stream indeed works effectively to boost the segmentation accuracy.  



\subsubsection{Performance of Camouflaged Object Segmentation}

It is interesting for the saliency community to see the segmentation capabilities of saliency systems to segment camouflaged objects (which should actually not be possible due to different properties of those objects).  Therefore, we show results of FCNs using pre-trained models for salient object segmentation in Table~\ref{tab:result}. Although results are moderate (the worst results among compared methods for each FCN), it also illustrates the potential ability to transfer from salient object segmentation to camouflaged object segmentation due to the ambiguous border of these kinds of object. This shows the possibility of using existing saliency models to tackle camouflaged object segmentation problem.

Table~\ref{tab:result} shows that the fine-tuned FCN models achieve around 60$+$\% accuracy in $F_\beta$ and IOU (around 0.1 in MAE).  Naturally, the model fine-tuned with the dataset used for testing performs better than the pre-trained model or the model fine-tuned with the other dataset. We also observe the model fine-tuned with CAMO-COCO performs better than the one fine-tuned with CAMO. This is because CAMO is a subset of CAMO-COCO. These mean that salient object segmentation methods are, to some extent, capable of segmenting even a not-target object such as a camouflaged object. This may come from the ambiguity of the differences between salient objects and camouflaged objects. This, simultaneously, also supports the use of an FCN developed for salient object segmentation as the segmentation stream in our ANet. 

We then draw our attention to the performances of ANet using an FCN. Since the segmentation stream of ANet is trained using CAMO and its output is multiplied by the output of the classification stream, the segmentation ability of ANet is limited by that of the employed FCN fine-tuned with CAMO. In other words, the performance of ANet on CAMO is expected to be comparable with that of the employed FCN (fine-tuned with CAMO), which is confirmed in Table~\ref{tab:result}. When we evaluate ANet on CAMO-COCO, however, the performance of ANet can be better than that of the employed FCN (fine-tuned with CAMO). This can be explained by the nature of CAMO-COCO where an image may not contain any camouflaged object, and thus the role of the classification stream becomes more crucial; in other words, the accuracy of the classification stream boosts up the segmentation accuracy of ANet.  The segmentation accuracy of the FCN, in contrast, is degraded because it is not trained on CAMO-COCO. A similar conclusion can be derived from the comparison between ANet and the employed FCN fine-tuned with CAMO-COCO.
Accordingly, ANet works with comparable accuracy on \textit{both} CAMO and CAMO-COCO while each FCN works well \textit{only on one} of the two datasets. Indeed, this can be observed in Table~\ref{tab:result}. This clearly demonstrates that ANet can maintain with flexibility a good segmentation accuracy, regardless whether a camouflaged object is present in the input image or not. 

We also statistically evaluated the significance of the differences using the t-test with the significance level of $90\%$. For CAMO-COCO dataset, we confirmed that on all the metrics, ANet significantly outperforms the other methods mostly or is at least similar to the best methods. For CAMO dataset, we also confirmed ANet is significantly better than FCNs fine-tuned on CAMO-COCO, and FCNs fine-tuned on CAMO are significantly better than ANet. This means ANet's ability for CAMO dataset is between FCNs fine-tuned on CAMO and FCNs fine-tune on CAMO-COCO. This is reasonable because the performance of the classification stream is not perfect, leading to slightly reducing the performance on CAMO.

To summarize, ANet has good performance for {\it both} CAMO and CAMO-COCO datasets. On the other hand, FCNs fine-tuned with CAMO-COCO have good performance for only CAMO-COCO but not for CAMO; whereas FCNs fine-tuned with CAMO have good performance for only CAMO but not for CAMO-COCO ({\it not for both}).

\subsubsection{Performance of Camouflaged Object Classification}

In order to evaluate the level of accuracy achieved by our classification stream, we compared the performance of our classification stream with that of the SVM-based classification using the Bag-of-Words model~\citep{FeiFei-CVPR2005} with SIFT features~\citep{Lowe-ICCV1999} (denoted by SVM-BoW) and CNNs, including AlexNet~\citep{Krizhevsky-NIPS2012} and VGG-16~\citep{Simonyan-ILSVRC2014}. For CNNs, we fine-tuned them on our CAMO-COCO dataset from pre-trained models on the ImageNet dataset~\citep{Russakovsky-IJCV2015}. The results are shown in Table~\ref{tab:accuracy_classification}, indicating that the accuracy of our classification stream achieves around 90\%, which outperforms SVM-BoW. ANet baseline models are also significantly better than both AlexNet and VGG-16. As a closer look, the classification stream achieves high accuracy on few training iterations (3 epochs).

\begin{table}[t]
\centering
\caption{The accuracy (\%) of camouflaged object classification on the CAMO-COCO dataset.}
\label{tab:accuracy_classification}
\small
\begin{tabular}{l|c}
\toprule
\textbf{Method} & \textbf{Accuracy} \\
\midrule
SVM-BoW~\citep{FeiFei-CVPR2005} & 54.0 \\
AlexNet~\citep{Krizhevsky-NIPS2012} & 76.0 \\
VGG-16~\citep{Simonyan-ILSVRC2014} & 88.6 \\
\midrule
ANet-DHS & 91.4 \\
ANet-DSS & 89.2 \\
ANet-SRM & 90.6 \\
ANet-WSS & 89.6 \\
\bottomrule
\end{tabular}
\end{table}

\subsubsection{Computational Efficiency}

We further evaluated the computational efficiency of all the baseline models. We also evaluated the average running time of the original FCN models employed in our ANet.  The results are shown in Table~\ref{tab:time}.  We observe that 
stacking the classification stream does not incur much more processing time, indicating that the ANet is able to keep the running time computationally efficient.

\begin{table}[t]
\centering
\caption{The average wall-clock time in millisecond for each image.}
\label{tab:time}
\small
\begin{tabular}{l|cc}
\toprule
\textbf{Method} & \textbf{FCN} & \textbf{ANet (baseline)} \\
\midrule
DHS~\citep{Liu-CVPR2016} & 63 & 73 \\
DSS~\citep{Hou-CVPR2017} & 76 & 79 \\
SRM~\citep{Wang-ICCV2017} & 81 & 86 \\
WSS~\citep{Wang-CVPR2017} & 61 & 65 \\
\bottomrule
\end{tabular}
\end{table}

\subsection{Discussion}

\subsubsection{\Edit{Failure Cases}}
\label{failure_cases}

\begin{figure*}[t]
    \centering
    \includegraphics[width=1\textwidth]{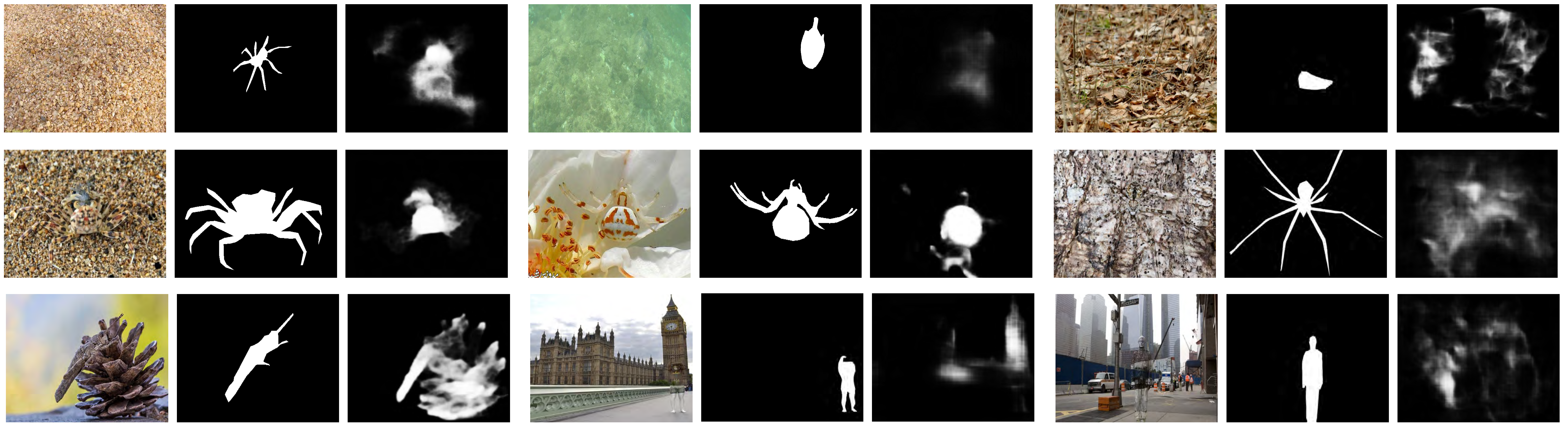}
    \caption{Failure cases of FCNs on CAMO. Each case is shown in a horizontal triplet of images, from left to right: input image, ground-truth, and result by the best FCN (fine-tuned with CAMO) in this order.}
\label{fig:failure}
\end{figure*}

\Edit{The accuracy of our baselines on CAMO for camouflaged object segmentation is far lower than the state-of-the-art accuracy for salient object segmentation, which can achieve up to round 90\% in F$_\beta$. This is mainly caused by the insufficient ability of the segmentation stream when we transfer domain from salient object segmentation to camouflaged object segmentation. Figure~\ref{fig:failure} shows some failure segmentation results obtained by our used FCNs where input images involve challenging scenarios in CAMO (object appearance and background clutter in the first row; the shape complexity of insect legs in the second row; distraction in the third row).}

\subsubsection{\Edit{Joint Training}}
\label{joint_training}

 The classification stream is expected to boost the segmentation accuracy in ANet, however, fusing the two streams by the multiplication is still insufficient.  Further effective fusion of the two streams should be explored to improve the segmentation accuracy. 


For a two-stream network, jointly training two streams after training each stream separately is known as a strategy to improve the performance~\citep{Jain-CVPR2017}.  Following this strategy, we jointly trained the ANet (baseline) using CAMO-COCO where we used 10 epochs (10k iterations) with the learning-rates of $10^{-4}$ (for the segmentation stream) and $10^{-6}$ (for the classification stream). Other parameters are set similarly to training two streams individually. The loss function for joint-training is defined as the summation of the loss of the classification stream and that of the segmentation stream: 
 \begin{equation}
 \label{eq:loss}
 \mathcal{L}=\mathcal{L}_{\rm seg}(\hat{y}_{\rm seg},y_{\rm seg})+\mathcal{L}_{\rm cls}(\hat{y}_{\rm cls},y_{\rm cls}),
 \end{equation}
where $\hat{y}_{\rm seg}$, $y_{\rm seg}$, $\hat{y}_{\rm cls}$, and $y_{\rm cls}$ denote outputs and ground-truth labels of the segmentation stream and the classification stream, respectively. $\mathcal{L}_{\rm cls}$ is the soft-max loss of the classification stream and $\mathcal{L}_{\rm seg}$ is the specific loss of the corresponding FCN implemented in the segmentation stream.

Table~\ref{tab:joint-training} illustrates the improvement of performances by introducing the joint-training to ANet. We see that differently from other two-stream network cases, jointly-training ANet does not lead to better performance for all methods. DHS and WSS reduce accuracy while DSS and SRM slightly increase accuracy (see Table~\ref{tab:joint-training}). This can be explained as follows. Each stream of ANet has its own task and thus using training data suitable for the task is mandatory to enhance its ability. During the joint-training of ANet, however, data used for training cannot be suitable for each of the tasks of the two streams, resulting in interfering with each other stream. This is also facilitated by the branch structure of ANet. Accordingly, joint-training potentially brings to ANet some collapse in improving training of each stream. 

Therefore, jointly training two streams may not bring gain to ANet due to opposite attributes of camouflaged objects and non-camouflaged objects. The two-step training approach is sufficient for ANet for the task of camouflaged object segmentation on the CAMO-COCO dataset. Further investigation on this issue is left for future work.

\begin{table*}[t]
\centering
\caption{Performance of joint-training on CAMO dataset and CAMO-COCO dataset. The best results are shown in \textcolor{blue}{\textbf{blue}}.}
\label{tab:joint-training}
\resizebox{1\textwidth}{!}{%
\begin{tabular}{|l|c|cc|cc|l|c|cc|cc|}
\toprule
 \textbf{Dataset in test} & \multicolumn{5}{c|}{\textbf{CAMO}} &  & \multicolumn{5}{c|}{\textbf{CAMO-COCO}} \\ \cline{1-6} \cline{8-12} 
\textbf{Method} & \textbf{} & \multicolumn{2}{c|}{\textbf{Adaptive Threshold}} & \multicolumn{2}{c|}{\textbf{Fixed Threshold}} &  & \textbf{} & \multicolumn{2}{c|}{\textbf{Adaptive Threshold}} & \multicolumn{2}{c|}{\textbf{Fixed Threshold}} \\
 & \textbf{MAE $\Downarrow$} & \textbf{F$_\beta$ $\Uparrow$} & \textbf{IOU $\Uparrow$} & \textbf{F$_\beta$ $\Uparrow$} & \textbf{IOU $\Uparrow$} &  & \textbf{MAE $\Downarrow$} & \textbf{F$_\beta$ $\Uparrow$} & \textbf{IOU $\Uparrow$} & \textbf{F$_\beta$ $\Uparrow$} & \textbf{IOU $\Uparrow$} \\ \cmidrule{1-6} \cmidrule{8-12} 
 
ANet-DHS (baseline) & \textcolor{blue}{0.130} & \textcolor{blue}{0.626} & \textcolor{blue}{0.437} & \textcolor{blue}{0.631} & \textcolor{blue}{0.423} &  & \textcolor{blue}{0.072} & \textcolor{blue}{0.812} & \textcolor{blue}{0.712} & \textcolor{blue}{0.814} & \textcolor{blue}{0.705} \\
ANet-DHS (+joint training) & 0.201 & 0.509 & 0.314 & 0.490 & 0.130 &  & 0.109 & 0.749 & 0.628 & 0.758 & 0.564 \\
\midrule 
ANet-DSS (baseline) & 0.132 & 0.587 & 0.404 & 0.607 & 0.390 &  & 0.067 & 0.795 & 0.701 & 0.804 & 0.694 \\
ANet-DSS (+joint training) & \textcolor{blue}{0.126} & \textcolor{blue}{0.644} & \textcolor{blue}{0.441} & \textcolor{blue}{0.651} & \textcolor{blue}{0.417} &  & \textcolor{blue}{0.064} & \textcolor{blue}{0.822} & \textcolor{blue}{0.719} & \textcolor{blue}{0.825} & \textcolor{blue}{0.707} \\
\midrule 
ANet-SRM (baseline) & 0.126 & 0.654 & 0.475 & 0.662 & \textcolor{blue}{0.466} &  & 0.069 & 0.826 & 0.732 & 0.830 & \textcolor{blue}{0.727} \\
ANet-SRM (+joint training) & \textcolor{blue}{0.123} & \textcolor{blue}{0.680} & \textcolor{blue}{0.481} & \textcolor{blue}{0.682} & 0.454 &  & \textcolor{blue}{0.063} & \textcolor{blue}{0.839} & \textcolor{blue}{0.733} & \textcolor{blue}{0.841} & 0.726 \\
\midrule 
ANet-WSS (baseline) & \textcolor{blue}{0.140} & \textcolor{blue}{0.661} & \textcolor{blue}{0.459} & \textcolor{blue}{0.643} & \textcolor{blue}{0.407} &  & 0.078 & \textcolor{blue}{0.826} & \textcolor{blue}{0.710} & \textcolor{blue}{0.820} & \textcolor{blue}{0.697} \\
ANet-WSS (+joint training) & 0.145 & 0.626 & 0.394 & 0.628 & 0.324 &  & \textcolor{blue}{0.074} & 0.813 & 0.692 & 0.816 & 0.662 \\
\bottomrule
\end{tabular}
}
\end{table*}


\section{Conclusion}\label{sec:conclusion}

\Edit{In this paper, we addressed an interesting yet challenging problem of camouflaged object segmentation by providing a new image dataset of camouflaged object segmentation where each image is manually annotated with pixel-wise ground-truth. We believe that our novel dataset will promote new advancements on camouflaged object segmentation. We also aim to explore camouflaged object segmentation on videos in the near future.}

\Edit{We proposed a simple and flexible end-to-end network, namely Anabranch Network, for camouflaged object segmentation where the classification stream and the segmentation stream are effectively combined to show the baseline performance. To show effectiveness of our proposed framework, which can boot the segmentation using classification, we applied it to different FCNs. Extensive experiments conducted on the newly built dataset demonstrate the superiority of the proposed network. In addition, our method is computationally efficient.}

\Edit{This paper focused on only regions, but the potential idea of utilizing classification into segmentation can be useful when applying for instance segmentation in appropriate ways. For examples, instance classification can be combined with instance segmentation, where each instance is classified whether it is camouflaged or salient. This is left for the future work.}


\section*{Acknowledgment}
This work is in part granted by University of Dayton SEED Grant (US), and by SOKENDAI Short-Stay Study Abroad Program (Japan). We also thank Vamshi Krishna Madaram and Zhe Huang (University of Dayton) for their support in the annotation of the CAMO dataset. We gratefully acknowledge NVIDIA for the support of GPU. 


\bibliographystyle{model2-names}
\bibliography{shortbib}

\end{document}